\documentclass[a4paper,fleqn]{cas-dc}

\usepackage[numbers, sort]{natbib}
\def\tsc#1{\csdef{#1}{\textsc{\lowercase{#1}}\xspace}}
\tsc{WGM}
\tsc{QE}
\tsc{EP}
\tsc{PMS}
\tsc{BEC}
\tsc{DE}
\usepackage{amsmath,graphicx,mathrsfs,booktabs,type1cm,amssymb,threeparttable,multirow, booktabs, changepage}

\usepackage{footnote}
\usepackage{color, xcolor, soul}
\usepackage{verbatim}
\usepackage{bm}
\usepackage{url}
\usepackage{array}
\usepackage{stfloats}
\usepackage{float}
\usepackage{graphicx}
\usepackage{makecell, rotating}
\usepackage{threeparttable}
\usepackage{textcomp}
\usepackage[caption=false, justification=centering]{subfig}
\usepackage{enumitem}

\usepackage{media9}

\begin{document}
\begin{sloppypar}
\let\WriteBookmarks\relax
\def\floatpagepagefraction{1}
\def\textpagefraction{.001}

\shorttitle{S2Dialog: Multimodal Dialogue Retrieval with Semantic and Acoustic-Style Modeling}

\shortauthors{Xueqi Wang et~al.}

\title [mode = title]{S2Dialog: Multimodal Dialogue Retrieval with Semantic and Acoustic-Style Modeling}                      



%
\author[1]{Xueqi Wang}[type=editor,
                        ]

\ead{xueqiwang7@163.com}

\credit{Conceptualization, Methodology, Formal analysis,
Data curation, Writing – original draft, Writing – review & editing, Validation, Funding acquisition.}

\affiliation[1]{organization={College of Computer Science, Inner Mongolia University},
    city={Hohhot},
    country={China}}

\affiliation[2]{organization={University of Electronic Science and Technology of China, Shenzhen Campus},
 city={Shenzhen}, country={China}}


\author[1]{Zhigang Wang}[style=chinese]
\credit{Methodology, Writing – original draft, Data curation, Resources}

\author[2]{Runqing Zhang}[]
\credit{Methodology}

\author[1]{Zhenqi Jia}[]
\credit{Methodology}

\author[1]{Junfeng Zhao}[]
\credit{Writing – review & editing, Validation, Funding acquisition.}

\nonumnote{Corresponding Author: Junfeng Zhao.}

\begin{abstract}
Multimodal dialogue retrieval aims to retrieve dialogues from multimodal dialogue banks that are similar to a target dialogue in terms of both textual semantics and acoustic conversational styles. Such dialogue-level retrieval is crucial for many dialogue-related tasks, including Emotion Recognition in Conversation, Spoken Dialogue Systems, and Conversational Speech Synthesis, where external dialogue examples can provide valuable semantic and stylistic references. However, existing retrieval methods are still largely limited to utterance-level or unimodal matching, and often fail to capture the global semantic coherence and stylistic consistency of an entire dialogue. To address this gap, we propose \textbf{S2Dialog}, a unified framework for dialogue-level semantic-style retrieval from multimodal dialogue banks. Specifically, S2Dialog consists of a Dialogue-level Textual Retriever and a Dialogue-level Acoustic Retriever, which encode the textual and acoustic modalities of a dialogue into dialogue-level representations, respectively. To further enhance multimodal retrieval, we introduce Dialogue-level Textual-Acoustic Contrastive Learning, which aligns semantically and stylistically similar dialogues while distinguishing unrelated ones. Extensive experiments on the multimodal dialogue dataset DailyTalk demonstrate that S2Dialog achieves outstanding retrieval performance. The code will be released at \url{https://github.com/anonymous-retrieval/S2Dialog}.
\end{abstract}

\begin{highlights}
\item We introduce \textbf{S2Dialog}, a dialogue-level semantic-style retrieval framework for multimodal dialogue banks.

\item We design dedicated textual and acoustic dialogue-level retrievers and introduce Dialogue-level Textual-Acoustic Contrastive Learning to jointly model dialogue semantics and conversational styles, enabling more discriminative semantic-style retrieval from complete multimodal dialogues.

\item Extensive experiments on DailyTalk demonstrate the effectiveness of S2Dialog over representative retrieval baselines.
\end{highlights}


\maketitle

\section{Introduction}
Multimodal Dialogue Retrieval (MDR) aims to model the textual and acoustic modalities of a target dialogue to retrieve dialogues from a Multimodal Dialogue Bank (MDB) that are similar in semantics or conversational style. With the rapid development of Human-Computer Interaction (HCI), dialogue tasks are playing an increasingly important role in various applications such as intelligent assistants, customer service systems, and voice-based interaction platforms \cite{zhou2020design, seaborn2021voice, mctear2022conversational}. To enhance the understanding and generation capabilities of dialogue systems, integrating external knowledge through MDR has become a key approach to improving the performance of dialogue-related tasks \cite{wang2023retrieval, wang2024retrieval, liu2025retrieval, jia2025intra, hu2025chain, liu2026emphasis, jia2025multimodal, jia2026auemochat}.

Retrieval mechanisms have become an important way to provide external context and knowledge for dialogue-centric systems \cite{lewis2020retrieval, yu2024evaluation, zhang2025amns, ram2023incontext, asai2023selfrag}. In recent years, retrieval-augmented methods have been increasingly explored in dialogue-related tasks. For example, DFA-RAG \cite{sun2024dfarag} retrieves historical dialogue examples to guide dialogue-state transitions; UniMS-RAG \cite{wang2024unims} leverages multi-source knowledge retrieval to maintain contextual and personalized response generation; ConvRAG \cite{ye2024boosting} refines user queries with dialogue history and retrieves fine-grained external evidence; and RADKA-CSS \cite{liu2025retrieval} improves conversational speech synthesis by retrieving dialogues that are similar to the target dialogue in terms of semantics and conversational style. These studies suggest that retrieving relevant dialogue knowledge can effectively enhance dialogue understanding and generation.

\begin{figure}
    \centering
    \includegraphics[width=1\linewidth]{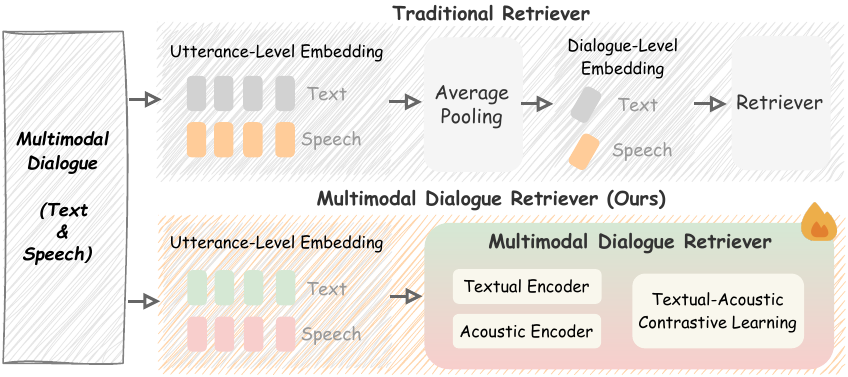}
    \caption{From traditional retriever to our proposed multimodal dialogue retriever.}
    \label{fig:idea}
    \vspace{-1em}
\end{figure}

Although dialogue retrieval has achieved significant progress in various dialogue tasks, existing approaches remain largely confined to utterance-level or unimodal perspectives. As illustrated in Fig. \ref{fig:idea}, these approaches typically rely on general-purpose pre-trained models to derive dialogue-level representations by simply aggregating utterance-level or unimodal features via average pooling \cite{su2023one}. These representations are not specifically optimized for dialogue retrieval and lack dedicated mechanisms for capturing complex dialogue-level and multimodal interactions. Consequently, these approaches struggle to effectively model the semantic coherence and stylistic consistency of entire conversations \cite{liu2024lost}, and fail to align semantics and styles across different modalities, thereby limiting their performance in multimodal dialogue retrieval.

To address the above gap, we propose \textbf{S2Dialog}, a retrieval framework designed for multimodal dialogue. S2Dialog consists of two dedicated dialogue-level retrievers: a Textual Retriever and an Acoustic Retriever. Specifically, the Textual Retriever employs a dialogue-level textual encoder to encode the textual modality of an entire dialogue into a dialogue-level semantic representation. Similarly, the Acoustic Retriever utilizes a dialogue-level acoustic encoder to capture the acoustic conversational style of the full dialogue and produce a dialogue-level style representation. To further bridge textual semantics and acoustic styles, we introduce Dialogue-level Textual-Acoustic Contrastive Learning, which pulls semantically and stylistically similar dialogues closer in the multimodal representation space while pushing dissimilar dialogues apart. This enables S2Dialog to learn more discriminative dialogue-level representations for semantic-style retrieval. In summary, the contributions of this paper are as follows:

\begin{itemize}
\item We introduce \textbf{S2Dialog}, a dialogue-level semantic-style retrieval framework for multimodal dialogue banks.

\item We design dedicated textual and acoustic dialogue-level retrievers and introduce Dialogue-level Textual-Acoustic Contrastive Learning to jointly model dialogue semantics and conversational styles, enabling more discriminative semantic-style retrieval from complete multimodal dialogues.

\item Extensive experiments on DailyTalk demonstrate the effectiveness of S2Dialog over representative retrieval baselines.
\end{itemize}

\section{Related Works}
\subsection{Dialogue Retrieval}
Dialogue retrieval aims to retrieve dialogues from an MDB that are semantically or stylistically relevant to a target conversation \cite{li2025dialogue,bartl2017retrieval}. Existing approaches mainly construct dialogue-level representations through pooling or summarization over dialogue histories. For example, UniRetriever \cite{wang2024uniretriever}, which dynamically aggregates query-relevant dialogue utterances through attention-based pooling; HAConvDR \cite{mo2024history}, which incorporates relevant historical turns for conversational retrieval; MARS \cite{mu2026hearing}, which performs cross-modal similarity retrieval over speech and text histories; and RADKA-CSS \cite{liu2025retrieval}, which jointly models semantic and speaking-style similarity in multimodal dialogues.

Different from these approaches, \textbf{S2Dialog} introduces dedicated textual and acoustic retrievers that explicitly model complete conversational sequences, enabling more effective dialogue-level semantic-style retrieval from MDB.

\begin{figure*}
    \centering
    \includegraphics[width=1\linewidth]{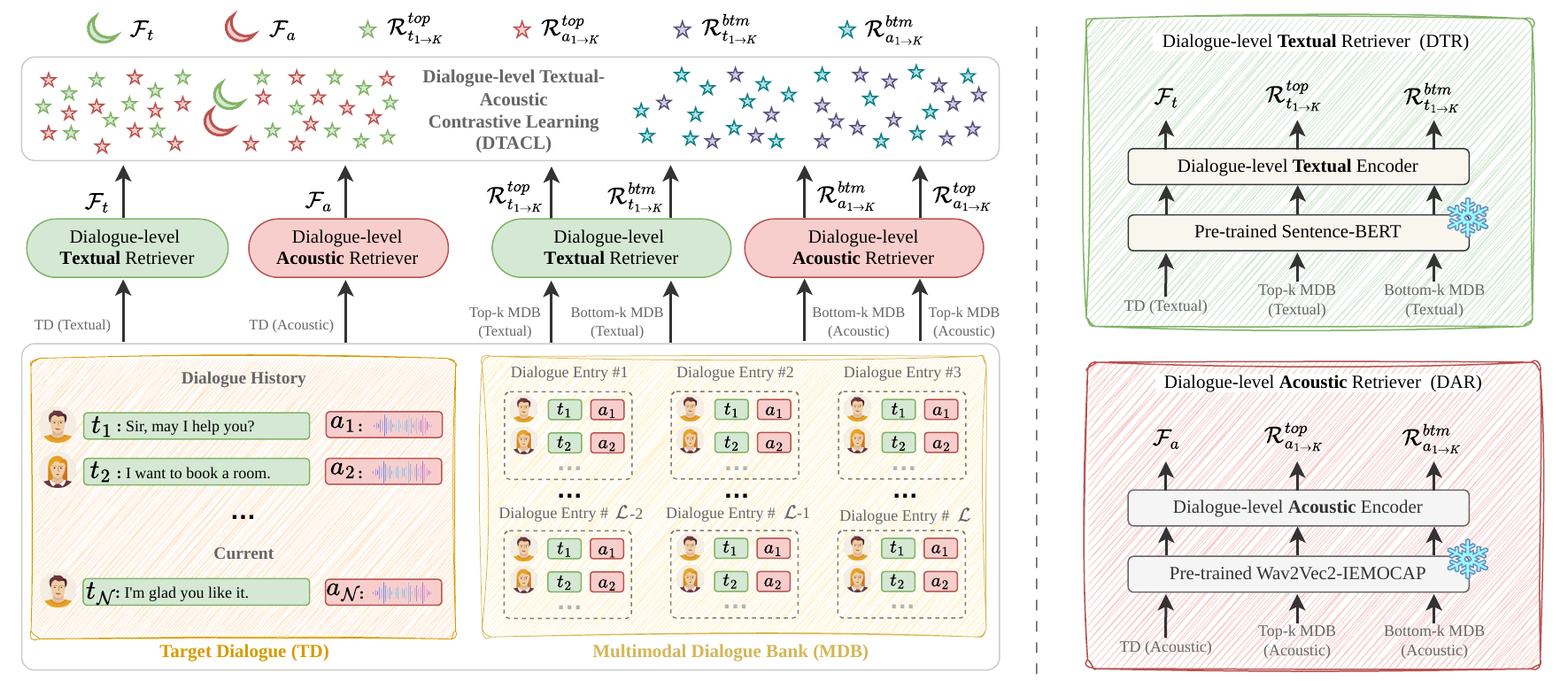}
    \caption{The overview of S2Dialog consists of Dialogue-level Textual Retriever, Dialogue-level Acoustic Retriever, and Dialogue-level Textual-Acoustic Contrastive Learning.}
    \label{fig:main}
\end{figure*}

\subsection{Contrastive Learning}
Contrastive learning aims to enhance the discriminative power of learned representations by pulling positive pairs closer and pushing dissimilar samples apart in a unified embedding space \cite{khosla2020supervised, le2020contrastive, tian2020makes, tan2023contrastive}. This paradigm has been widely extended to cross-modal representation learning, where heterogeneous modalities are aligned within a shared latent space. For example, CLIP \cite{radford2021learning} jointly trains image and text encoders to align visual and textual concepts, enabling robust zero-shot generalization. StyleDiffusion \cite{wang2023stylediffusion} integrates contrastive representation learning with conditional generation to disentangle stylistic attributes from semantic content. Similarly, CLAP \cite{ye2023clapspeech} aligns audio and text representations through contrastive objectives, while CALM \cite{meng2023calm} optimizes the correlation between speaker style embeddings and style-related textual features to improve expressive speech synthesis.

In our work, we extend cross-modal contrastive learning to dialogue-level semantic-style retrieval. Specifically, we introduce a Dialogue-level Textual-Acoustic Contrastive Learning module to bridge textual semantics and acoustic conversational styles across complete dialogues. By optimizing dialogue-level textual and acoustic representations in a unified retrieval space, the proposed module encourages cross-modal alignment and improves the discriminative capacity of S2Dialog for retrieving semantically and stylistically similar dialogues from MDB.

\section{Methodology}
\subsection{Model Overview}
As shown in Fig. \ref{fig:main}, the proposed \textbf{S2Dialog} consists of three components: (1) Dialogue-level Textual Retriever (DTR), (2) Dialogue-level Acoustic Retriever (DAR), and (3) Dialogue-level Textual-Acoustic Contrastive Learning module (DTACL). DTR and DAR encode the textual and acoustic modalities of an input dialogue into dialogue-level semantic and style representations, respectively. DTACL is designed to align semantically and stylistically similar dialogues while separating unrelated ones in the multimodal representation space, thereby improving dialogue-level semantic-style retrieval from MDB.

\subsection{Dialogue-level Textual Retriever (DTR)}
The Dialogue-level Textual Retriever (DTR) is designed to project dialogue texts of varying lengths into a unified, fixed-dimensional semantic space. To model the progression from local utterance-level semantics to global dialogue-level semantics, DTR adopts a two-stage representation learning process.

\textbf{1) Utterance-level Semantic Extraction:} Given a target dialogue $TD$ consisting of $N$ utterances $\{t_1, t_2, \dots, t_N\}$, we first employ a pre-trained Sentence-BERT\footnote{\label{sentencebert}https://huggingface.co/sentence-transformers/distiluse-base-multilingual-cased-v1} \cite{reimers-gurevych-2019-sentence} as a frozen backbone to extract utterance-level semantic embeddings. Each utterance $t_i$ is transformed into a high-dimensional vector $\mathbf{e}_{t_i} \in \mathbb{R}^{d_{sent}}$, where $d_{sent}=512$. This step preserves the linguistic semantics of each dialogue turn before dialogue-level aggregation.

\textbf{2) Textual Temporal Contextual Aggregation:} To capture long-range dependencies and the evolving semantic flow within a dialogue, we utilize a Gated Recurrent Unit (GRU) as the dialogue-level textual encoder. The sequence of utterance embeddings $\{\mathbf{e}_{t_1}, \dots, \mathbf{e}_{t_N}\}$ is fed into the GRU to compute hidden state representations:
\begin{equation}
    \mathbf{h}_{t_i} = \text{GRU}_{text}(\mathbf{e}_{t_i}, \mathbf{h}_{t_{i-1}}),
\end{equation}
where $\mathbf{h}_{t_i}$ denotes the hidden state at the $i$-th turn. The final hidden state $\mathbf{h}_{t_N}$ is regarded as the dialogue-level summary of the textual discourse. To further refine this representation for retrieval, we pass $\mathbf{h}_{t_N}$ through a non-linear projection head consisting of a multi-layer perceptron (MLP) with ReLU activation, yielding the final dialogue-level semantic representation $\mathcal{F}_t$ for the target dialogue.

To construct a unified retrieval space, DTR is also used to encode candidate dialogues from the MDB. Specifically, the Top-K similar and Bottom-K dissimilar dialogues are mapped into their textual representation sets $\mathcal{R}^{top}_{t_{1 \rightarrow K}}$ and $\mathcal{R}^{btm}_{t_{1 \rightarrow K}}$.

\subsection{Dialogue-level Acoustic Retriever (DAR)}
Parallel to the textual branch, the Dialogue-level Acoustic Retriever (DAR) is designed to project acoustic features into a unified, fixed-dimensional style space. To model the temporal dynamics of speech across a conversation, DAR adopts a two-stage representation learning process.

\textbf{1) Utterance-level Acoustic Extraction:} For the acoustic modality, we leverage a pre-trained Wav2Vec2-IEMOCAP as the frozen feature extractor. Given a target dialogue consisting of $N$ speech segments $\{a_1, a_2, \dots, a_N\}$, the encoder generates a sequence of utterance-level acoustic representations. Each speech segment $a_i$ is transformed into a high-dimensional vector $\mathbf{e}_{a_i} \in \mathbb{R}^{d_{acou}}$, where $d_{acou}=768$. This stage captures phonetic and prosodic cues that reflect the speaker's delivery within each dialogue turn.

\textbf{2) Acoustic Temporal Contextual Aggregation:} To aggregate utterance-level acoustic representations into a coherent dialogue-level descriptor, we employ a GRU as the dialogue-level acoustic encoder. This module captures stylistic consistency and acoustic transitions throughout the dialogue. The sequence of acoustic embeddings $\{\mathbf{e}_{a_1}, \dots, \mathbf{e}_{a_N}\}$ is fed into the GRU to compute hidden state representations:
\begin{equation}
    \mathbf{h}_{a_i} = \text{GRU}_{acou}(\mathbf{e}_{a_i}, \mathbf{h}_{a_{i-1}}),
\end{equation}
where $\mathbf{h}_{a_i}$ denotes the hidden state at the $i$-th speech segment. The final hidden state $\mathbf{h}_{a_N}$ is regarded as the dialogue-level representation of the acoustic dynamics. To further refine this representation for retrieval, we pass $\mathbf{h}_{a_N}$ through a non-linear projection head consisting of an MLP with ReLU activation, yielding the final dialogue-level style representation $\mathcal{F}_a$ for the target dialogue.

Similarly, DAR encodes the Top-K similar and Bottom-K dissimilar candidate dialogues from the MDB into acoustic representation sets $\mathcal{R}^{top}_{a_{1 \rightarrow K}}$ and $\mathcal{R}^{btm}_{a_{1 \rightarrow K}}$, respectively.

\subsection{Dialogue-level Textual-Acoustic Contrastive Learning (DTACL)}
To enhance dialogue-level semantic-style retrieval, we propose Dialogue-level Textual-Acoustic Contrastive Learning (DTACL), which learns a unified retrieval space by pulling dialogues with similar textual semantics and acoustic styles closer while pushing unrelated dialogues apart. Unlike conventional contrastive learning that focuses on utterance-level or unimodal representations, DTACL performs contrastive optimization over complete dialogues by jointly modeling textual and acoustic representations.

Given a target dialogue $TD$, the Dialogue-level Textual Retriever and Dialogue-level Acoustic Retriever produce its textual and acoustic representations, denoted as $\mathcal{F}_t$ and $\mathcal{F}_a$, respectively. For candidate dialogues from the MDB, we denote the textual and acoustic representations of the Top-K similar dialogues as $\mathcal{R}^{+}_{t}=\{\mathbf{r}^{+,k}_{t}\}_{k=1}^{K}$ and $\mathcal{R}^{+}_{a}=\{\mathbf{r}^{+,k}_{a}\}_{k=1}^{K}$, respectively. Similarly, the textual and acoustic representations of the Bottom-K dissimilar dialogues are denoted as $\mathcal{R}^{-}_{t}=\{\mathbf{r}^{-,k}_{t}\}_{k=1}^{K}$ and $\mathcal{R}^{-}_{a}=\{\mathbf{r}^{-,k}_{a}\}_{k=1}^{K}$.

For the textual representation $\mathcal{F}_t$ of the target dialogue, the positive set consists of the acoustic representation of the same target dialogue and the textual-acoustic representations of Top-K similar dialogues:
\begin{equation}
\mathcal{P}_{t}
=
\{\mathcal{F}_a\}
\cup
\mathcal{R}^{+}_{t}
\cup
\mathcal{R}^{+}_{a}.
\end{equation}
Similarly, for the acoustic representation $\mathcal{F}_a$ of the target dialogue, the positive set is defined as:
\begin{equation}
\mathcal{P}_{a}
=
\{\mathcal{F}_t\}
\cup
\mathcal{R}^{+}_{t}
\cup
\mathcal{R}^{+}_{a}.
\end{equation}
The negative set is shared by both textual and acoustic anchors:
\begin{equation}
\mathcal{N}
=
\mathcal{R}^{-}_{t}
\cup
\mathcal{R}^{-}_{a}.
\end{equation}

Based on these sets, we define two contrastive losses. For the textual anchor $\mathcal{F}_t$, the contrastive loss is formulated as:
\begin{equation}
\mathcal{L}^{cl}_{t}
=
-\log
\frac{
\sum\limits_{\mathbf{u} \in \mathcal{P}_{t}}
\exp(\mathrm{sim}(\mathcal{F}_{t}, \mathbf{u}) / \tau)
}{
\sum\limits_{\mathbf{u} \in \mathcal{P}_{t} \cup \mathcal{N}}
\exp(\mathrm{sim}(\mathcal{F}_{t}, \mathbf{u}) / \tau)
},
\end{equation}
where $\mathrm{sim}(\cdot,\cdot)$ denotes cosine similarity and $\tau$ is a temperature parameter. Similarly, for the acoustic anchor $\mathcal{F}_a$, the contrastive loss is defined as:
\begin{equation}
\mathcal{L}^{cl}_{a}
=
-\log
\frac{
\sum\limits_{\mathbf{u} \in \mathcal{P}_{a}}
\exp(\mathrm{sim}(\mathcal{F}_{a}, \mathbf{u}) / \tau)
}{
\sum\limits_{\mathbf{u} \in \mathcal{P}_{a} \cup \mathcal{N}}
\exp(\mathrm{sim}(\mathcal{F}_{a}, \mathbf{u}) / \tau)
}.
\end{equation}
The final DTACL objective is:
\begin{equation}
\mathcal{L}^{cl}
=
\mathcal{L}^{cl}_{t}
+
\mathcal{L}^{cl}_{a}.
\end{equation}

By minimizing $\mathcal{L}^{cl}$, DTACL encourages textual semantics and acoustic conversational styles to be aligned at the dialogue level. Meanwhile, semantically and stylistically similar dialogues are pulled closer in the shared retrieval space, while unrelated dialogues are separated. This optimization enables S2Dialog to learn more discriminative dialogue-level representations for semantic-style retrieval from multimodal dialogue banks.

\section{Experiments}
\subsection{Dataset}

We evaluate S2Dialog on the DailyTalk \cite{lee2023dailytalk} dataset, a high-quality multimodal dialogue corpus. The dataset consists of 2,541 conversations comprising 23,773 audio clips, totaling 20 hours of audio data. Each conversation contains an average of 9.356 turns, with an average clip length of 3.282 seconds, providing sufficient context for dialogue-level modeling. The speech samples were recorded simultaneously by a male and a female speaker with a balanced distribution of utterances. To maintain high acoustic fidelity for retrieval, all samples are recorded at a sampling rate of 44.10 kHz and encoded in 16-bit format. Following standard practice, we partition the data into training, validation, and test sets in an 8:1:1 ratio.

\textbf{Top-K and Bottom-K Ground-Truth Construction.}
To construct reliable ground-truth (GT) labels for dialogue-level semantic-style retrieval, we adopt a multi-stage pipeline that combines automated cross-modal scoring with human refinement. \textbf{1) Dialogue-level Textual Similarity:}
Given the long-form nature of dialogues, we first employ \textit{BART-LARGE-CNN-SAMSUM}\footnote{\label{samsum}https://huggingface.co/philschmid/bart-large-cnn-samsum} to generate a summary of each dialogue. The summary is then embedded into a high-dimensional semantic space using \textit{Sentence-BERT} \cite{reimers-gurevych-2019-sentence}, and the textual similarity between dialogue pairs is measured by cosine similarity. \textbf{2) Dialogue-level Acoustic Similarity:}
For the acoustic modality, we use \textit{Wav2Vec2-IEMOCAP}\footnote{\label{wav2vec2iemocap}https://huggingface.co/speechbrain/emotion-recognition-wav2vec2-IEMOCAP} to extract utterance-level acoustic embeddings from each speech segment. To measure dialogue-level stylistic similarity, we compute the pairwise similarity matrix between utterances from two dialogues and use the average score as the global acoustic similarity. \textbf{3) Weighted Fusion and Human Refinement:}
We integrate textual and acoustic similarity scores through weighted fusion to obtain an overall semantic-style similarity score. Based on the fused scores, we initially select the Top-50 and Bottom-50 candidate dialogues for each target dialogue. To ensure that the final labels reflect real conversational similarity, three annotators familiar with the DailyTalk dataset manually re-rank these candidates. Disagreements are resolved by majority voting, resulting in consensus-based ground-truth retrieval labels.

\subsection{Implementation Details}
In S2Dialog, the Dialogue-level Textual Retriever consists of a pre-trained Sentence-BERT and a dialogue-level textual encoder. The dialogue-level textual encoder contains a GRU with input and hidden dimensions of 512, followed by a projection head composed of a linear layer from 512 to 256, a ReLU activation function \cite{agarap2018deep}, and another linear layer from 256 to 256. The Dialogue-level Acoustic Retriever consists of a pre-trained Wav2Vec2-IEMOCAP and a dialogue-level acoustic encoder. The dialogue-level acoustic encoder contains a GRU with input and hidden dimensions of 768, followed by a projection head composed of a linear layer from 768 to 256, a ReLU activation function, and another linear layer from 256 to 256. We train S2Dialog on a single A100 GPU with a batch size of 32 for 10 epochs.

\subsection{Evaluation Metrics}
To evaluate dialogue-level semantic-style retrieval, we adopt Recall \cite{zhu2004recall} as the main metric. Specifically, we report Recall@10, Recall@20, Recall@30, Recall@40, and Recall@50 to measure whether the retrieved results are consistent with the GT Top-K indices under different retrieval depths.

In addition, we report TopT@K, BtmT@K, TopA@K, and BtmA@K to evaluate the discriminative quality of the learned textual and acoustic representation spaces. These metrics measure the cosine similarities between the target dialogue and the corresponding Top-K or Bottom-K dialogues in the textual and acoustic latent spaces. We report them for $K \in \{1,25,50\}$.

\subsection{Baseline and Ablation Strategy}

To comprehensively evaluate the effectiveness of the proposed method, we implement four categories of representative dialogue-context modeling baselines. Note that we only adopt their dialogue representation, retrieval, or summarization components, rather than their original end-to-end task settings. Detailed descriptions of the four categories and their corresponding comparative models are provided below.

\noindent
1) \textbf{Text Pooling:} This category aggregates dialogue history through unimodal retrieval or similarity-based context pooling, including UIMTH \cite{xie2026uimth}, UniRetriever \cite{wang2024uniretriever}, and HAConvDR \cite{mo2024history}.
\begin{itemize}
    \item UIMTH \cite{xie2026uimth} is a text-based conversational retrieval method that enhances dialogue representations with TextGNN. It models temporal, semantic, and syntactic dependencies in dialogue text and aggregates unimodal textual features to improve fine-grained intent retrieval.

    \item UniRetriever \cite{wang2024uniretriever} performs conversational retrieval by dynamically selecting query-relevant dialogue utterances and aggregating them into a unified textual dialogue representation through attention-based pooling.

    \item HAConvDR \cite{mo2024history} is a conversational dense retrieval framework that reformulates the current query by selectively incorporating relevant historical turns through context denoising. It aggregates dialogue history into unified textual representations and further enhances retrieval via history-aware contrastive learning.
\end{itemize}

\noindent
2) \textbf{Multimodal Pooling:} This category performs cross-modal retrieval and similarity fusion over multimodal dialogue history, including MARS \cite{mu2026hearing}.
\begin{itemize}
    \item MARS \cite{mu2026hearing} is a retrieval-augmented conversational ASR framework that performs multimodal similarity pooling over speech and text histories. It retrieves Top-K acoustic and semantic neighbors separately, and then integrates cross-modal similarity scores through a ranking-based selection strategy to identify the most relevant historical context.
\end{itemize}

\noindent
3) \textbf{Text Summarization:} This category compresses dialogue history into condensed textual memory representations, including CONVERSE \cite{lee2024effective} and CORAL \cite{cheng2025coral}.
\begin{itemize}
    \item CONVERSE \cite{lee2024effective} is a summary-based conversational retrieval method that compresses long dialogue contexts into intent-oriented textual summaries and retrieves relevant dialogues through summary-aware embedding matching.

    \item CORAL \cite{cheng2025coral} is a conversational retrieval framework for multi-turn dialogue settings that employs summary-based conversation compression to improve retrieval efficiency and robustness under long-context scenarios.
\end{itemize}

\noindent
4) \textbf{Multimodal Summarization:} This category summarizes multimodal dialogue interactions into unified contextual representations, including RADKA-CSS \cite{liu2025retrieval}.
\begin{itemize}
    \item \textbf{RADKA-CSS} \cite{liu2025retrieval} adopts a multimodal summary-based retrieval strategy, where dialogue text histories are summarized into semantic embeddings for semantic retrieval, while dialogue audio histories are encoded into style embeddings for speaking-style retrieval. Historical dialogues are retrieved based on combined semantic and style similarity.
\end{itemize}

\begin{table*}[t!]
\caption{Recall Performance of Different Dialogue Retrieval Methods.}
\centering
\resizebox{0.77\linewidth}{!}{
\begin{tabular}{lccccc}
\toprule
\textbf{Systems}  & \textbf{Recall@10} & \textbf{Recall@20} & \textbf{Recall@30} & \textbf{Recall@40} & \textbf{Recall@50}\\
\midrule
Text Pooling  & 11.60    & 22.00    & 31.20   & 40.00   & 53.20    \\
Multimodal Pooling  & 21.20    & 35.20    & 50.00   & 60.80   & 68.40    \\
Text Summarization  & 16.40    & 28.00   & 38.80   & 45.60   & 54.00   \\
Multimodal Summarization  & 25.60    & 40.40    & 50.80   & 61.20   & 68.40    \\

\midrule
\textbf{S2Dialog} & \textbf{50.68}    & \textbf{63.01}    & \textbf{72.60}   & \textbf{77.17}   & \textbf{83.56}   \\
\bottomrule
\end{tabular}

}
\label{tab1}
\end{table*}
To verify the contribution of the different technical components in S2Dialog, we perform ablation studies by systematically removing core modules and loss functions: \textbf{\textit{Abl.1 w/o Textual Retriever}}: We remove the textual branch, relying solely on the Acoustic Retriever to optimize the training process. This is designed to verify the Textual Retriever’s capacity to provide a foundational semantic anchor. \textbf{\textit{Abl.2 w/o Acoustic Retriever}}: By removing the acoustic branch, the model is trained only via the Textual Retriever. This aims to validate the necessity of Acoustic Retriever in capturing stylistic nuances. \textbf{\textit{Abl.3 w/o $\mathcal{L}^{cl}_{t}$}}: The textual contrastive loss is removed, eliminating the explicit constraint for semantic alignment. This variant is intended to verify the role of $\mathcal{L}^{cl}_{t}$ in refining the textual latent space and ensuring that semantically related dialogues are pulled closer together. \textbf{\textit{Abl.4 w/o $\mathcal{L}^{cl}_a$}}: This variant removes the acoustic contrastive loss to assess the impact of acoustic contrastive optimization on retrieval performance. \textbf{\textit{Abl.5 w/o Bottom-K}}: We remove the Bottom-K dissimilar dialogues during the training phase. This configuration is designed to investigate the importance of hard negative mining in sharpening the discriminative boundary of the multimodal latent space and preventing representation collapse.

\begin{table*}[t!]
\caption{Recall Performance of S2Dialog and Ablation Variants.}
\centering
\resizebox{0.82\linewidth}{!}{
\begin{tabular}{lccccc}
\toprule
\textbf{Systems}  & \textbf{Recall@10} & \textbf{Recall@20} & \textbf{Recall@30} & \textbf{Recall@40} & \textbf{Recall@50}\\
\midrule

\hspace{3mm} \textit{Abl.1 w/o Textual Retriever}  & 39.60    & 54.80    & 69.60   & 75.13   & 79.60    \\
\hspace{3mm} \textit{Abl.2 w/o Acoustic Retriever}  & 47.60    & 62.80    & 71.80    & 75.20   & 80.80    \\
\hspace{3mm} \textit{Abl.3 w/o $\mathcal{L}^{cl}_{t}$}   & 45.32    & 59.75    & 70.24    & 73.31   & 79.57   \\
\hspace{3mm} \textit{Abl.4 w/o $\mathcal{L}^{cl}_a$}  & 42.19    & 57.73    & 70.47   & 73.28   & 77.58    \\
\hspace{3mm} \textit{Abl.5 w/o Bottom-K}   & 21.12    & 32.74    & 46.85   & 57.76   & 68.22    \\

\midrule

\textbf{S2Dialog} & \textbf{50.68}    & \textbf{63.01}    & \textbf{72.60}   & \textbf{77.17}   & \textbf{83.56}   \\
\bottomrule
\end{tabular}

}
\label{tab2}
\end{table*}

\section{Results and Discussion}
In this section, we present the experimental results of S2Dialog. We first compare S2Dialog with representative baseline methods. We then conduct ablation studies to analyze the contribution of different components. Finally, we examine similarity score trends across different retrieval ranks under various ablation settings to further understand the quality of the learned representation space. 

\subsection{Main Results}
To assess the effectiveness of S2Dialog in multimodal dialogue retrieval, we conduct a systematic evaluation against four representative baseline categories commonly used in dialogue tasks: Text Pooling, Multimodal Pooling, Text Summarization, and Multimodal Summarization.

As reported in Table \ref{tab1}, Text Pooling exhibits the lowest retrieval performance across all Recall metrics, reflecting the limitations of unimodal aggregation and simple mean-pooling, which tend to blur fine-grained semantic details and neglect conversational style. Incorporating acoustic information through Multimodal Pooling leads to a substantial performance gain, demonstrating that leveraging cross-modal context enhances dialogue-level representation. Text Summarization, by compressing dialogue histories into semantic embeddings, also improves retrieval relative to Text Pooling, indicating that structured compression can partially mitigate the information loss inherent in pooling. Among the baselines, Multimodal Summarization achieves the highest results, suggesting that combining semantic compression with multimodal integration provides the most robust foundation for dialogue-level retrieval under conventional strategies.

Despite these improvements, our proposed S2Dialog consistently outperforms all baselines across every evaluation metric, achieving Recall@10 of 50.68\% and Recall@50 of 83.56\%. This stable and substantial gain can be attributed to two key factors. First, the dedicated dialogue-level Textual and Acoustic Retrievers capture global semantic coherence and conversational style, avoiding the information loss inherent in pooling or summary-only approaches. Second, the Dialogue-level Textual-Acoustic Contrastive Learning explicitly aligns semantic and stylistic representations across modalities, enhancing the discriminative power of the learned latent space. Collectively, these components enable S2Dialog to model complex multimodal dependencies more effectively, resulting in higher-fidelity representations and more accurate retrieval of semantically and stylistically similar dialogues.

\subsection{Ablation Results}
To further analyze the contribution of different components in S2Dialog, we conduct ablation experiments by separately removing the textual retriever, the acoustic retriever, the contrastive objectives, and the Bottom-K negative samples. The results are reported in Table \ref{tab2}. Overall, all ablated variants underperform the full model across Recall@K metrics. The detailed analysis is as follows.

\textbf{\textit{Impact of Textual and Acoustic Retrievers:}} Removing either the textual or acoustic retriever leads to a clear performance decrease. Specifically, without the Textual Retriever, Recall@10 drops from 50.68\% to 39.60\%, while removing the Acoustic Retriever results in a smaller but still noticeable decrease to 47.60\%. This suggests that textual semantics play a central role in identifying semantically relevant dialogues, whereas acoustic representations provide complementary style-related information. The larger degradation caused by removing the Textual Retriever also indicates that semantic matching remains the primary signal in the current retrieval setting, while acoustic style modeling further refines dialogue-level matching.

\textbf{\textit{Role of Contrastive Learning:}} The removal of either textual or acoustic contrastive loss also weakens retrieval performance. When $\mathcal{L}^{cl}_{t}$ is removed, Recall@10 decreases to 45.32\%; when $\mathcal{L}^{cl}_{a}$ is removed, Recall@10 further decreases to 42.19\% and Recall@50 drops from 83.56\% to 77.58\%. These results suggest that the two contrastive objectives help regularize the textual and acoustic representation spaces from different perspectives. In particular, the relatively larger drop caused by removing $\mathcal{L}^{cl}_{a}$ implies that explicit acoustic contrastive supervision is beneficial for preserving style-sensitive distinctions in multimodal dialogue retrieval.

\begin{figure*}
    \centering
    \includegraphics[width=0.9\linewidth]{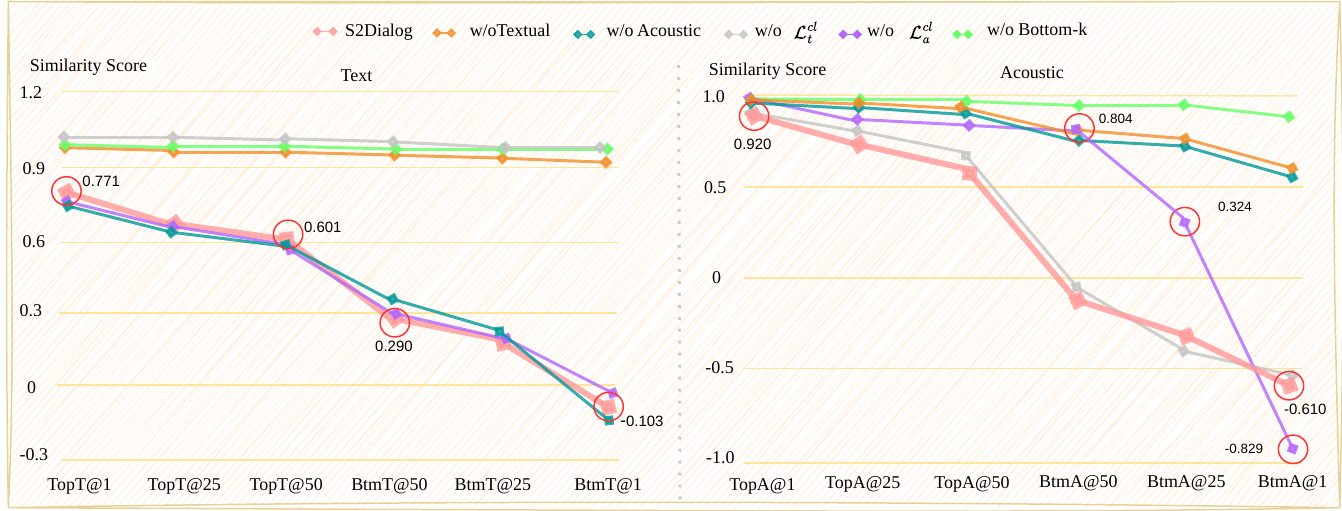}
    \caption{Similarity Score Trends under Different Ablation Settings Across Retrieval Ranks.}
    \vspace{-1em}
    \label{fig:an}
\end{figure*}

\textbf{\textit{Effect of Bottom-K Negative Samples:}} The most substantial degradation is observed when Bottom-K samples are excluded from training. Recall@10 decreases sharply from 50.68\% to 21.12\%, and Recall@50 decreases from 83.56\% to 68.22\%. This indicates that dissimilar dialogue samples provide important negative supervision for shaping the retrieval space. Without such negative signals, the model may still learn to bring related dialogues closer, but it becomes less capable of separating irrelevant or weakly related dialogues. Therefore, Bottom-K supervision appears to be important for learning a more discriminative dialogue-level semantic-style representation.

\begin{table}[t!]
\caption{Recall Performance of Multimodal Large Models and S2Dialog.}
\centering
\resizebox{1\linewidth}{!}{
\begin{tabular}{lccccc}
\toprule
\textbf{Systems} & \textbf{Recall@10} & \textbf{Recall@20} & \textbf{Recall@30} & \textbf{Recall@40} & \textbf{Recall@50} \\
\midrule
Qwen3-Omni 
& 16.45 & 27.27 & 41.23 & 55.02 & 63.38 \\

Qwen2.5-Omni 
& 14.29 & 25.54 & 37.66 & 47.62 & 54.11 \\

Kimi-Audio 
& 9.09 & 15.58 & 21.65 & 29.00 & 37.23 \\

Step-Audio 
& 16.02 & 30.30 & 43.72 & 51.95 & 60.17 \\
\midrule
\textbf{S2Dialog} 
& \textbf{50.68} & \textbf{63.01} & \textbf{72.60} & \textbf{77.17} & \textbf{83.56} \\
\bottomrule
\end{tabular}
}
\label{tab:large_model}
\end{table}

\subsection{Analysis of Multimodal Large Models}
We further compare S2Dialog with several representative multimodal large models, including Qwen3-Omni \cite{xu2025qwen3}, Qwen2.5-Omni \cite{xu2025qwen25omnitechnicalreport}, Kimi-Audio \cite{ding2025kimi}, and Step-Audio \cite{huang2025stepaudiounifiedunderstandinggeneration}. For these models, the complete multimodal dialogue is provided as input, and the final-layer hidden representation is used as the dialogue-level representation for retrieval. The results are shown in Table \ref{tab:large_model}.

Among the large-model baselines, Qwen3-Omni achieves the best Recall@10, Recall@40, and Recall@50, while Step-Audio performs best at Recall@20 and Recall@30. Nevertheless, all four models remain substantially below S2Dialog across retrieval performance. These results indicate that general-purpose multimodal models can encode useful textual and acoustic information from complete dialogues, but their representations are not specifically optimized for dialogue-level semantic-style retrieval. In contrast, S2Dialog explicitly models dialogue-level representation and optimizes the representation space through textual-acoustic contrastive learning, resulting in more discriminative representations for retrieval.

\subsection{Analysis of Similarity Trends Across Retrieval Ranks}
Fig. \ref{fig:an} further analyzes the quality of the learned representation space by tracking similarity scores across different retrieval ranks, from Top-K relevant dialogues to Bottom-K less related dialogues. Ideally, an effective dialogue-level representation should assign higher similarity scores to relevant dialogues and lower scores to unrelated ones, forming a clear descending trend across retrieval ranks.

\textbf{\textit{Textual Semantic Stability:}} 
For the textual modality (Fig. \ref{fig:an}, Left), the full S2Dialog exhibits a consistent decreasing trend, with the similarity score dropping from 0.771 at TopT@1 to -0.103 at BtmT@1. This indicates that the learned textual representation can better distinguish semantically relevant dialogues from less related ones. In contrast, removing the Textual Retriever or the textual contrastive objective $\mathcal{L}^{cl}_{t}$ leads to substantially flatter curves, where similarity scores remain high even for Bottom-K samples. This suggests that both the textual retriever and the textual contrastive loss contribute to shaping a more discriminative semantic space.

\textbf{\textit{Acoustic Style Sensitivity:}} 
For the acoustic modality (Fig. \ref{fig:an}, Right), the full model also shows a clear separation between Top-K and Bottom-K dialogues, with the similarity score decreasing from 0.920 at TopA@1 to -0.610 at BtmA@1. This trend suggests that S2Dialog is able to capture style-related differences between acoustically similar and dissimilar dialogues. By comparison, the variants without the Acoustic Retriever or Bottom-K samples maintain high similarity scores for bottom-ranked dialogues, indicating weaker separation in the acoustic representation space. These results show that acoustic modeling and negative-sample supervision are important for learning style-aware dialogue-level representations.

\textbf{\textit{Effect of Acoustic Contrastive Learning:}} 
The variant without $\mathcal{L}^{cl}_{a}$ presents a less stable similarity pattern in the acoustic space. Although it preserves high similarity scores for Top-K samples, the scores remain relatively high for several Bottom-K ranks and drop sharply only at the most dissimilar rank, reaching -0.829 at BtmA@1. This suggests that without explicit acoustic contrastive supervision, the acoustic representation space may be less well calibrated for ranking dialogues according to stylistic similarity. In other words, $\mathcal{L}^{cl}_{a}$ helps produce a smoother and more reliable similarity gradient across retrieval ranks.

Overall, these observations provide complementary evidence to the recall-based ablation results. The dedicated textual and acoustic retrievers help capture modality-specific dialogue-level information, while contrastive learning and Bottom-K supervision further improve the discriminability of the learned representation space. This enables S2Dialog to better model semantic relevance and stylistic consistency for multimodal dialogue retrieval.

\subsection{Analysis of Dialogue Temporal Modeling}
We further evaluate reversed, randomly shuffled, and bidirectional dialogue modeling. The results are reported in Table \ref{tab:temporal}. In particular, random shuffling reduces Recall@10 from 50.68\% to 37.66\%, while reversed modeling decreases Recall@10 and Recall@20 by 9.99 and 5.00 percentage points, respectively. These results indicate that S2Dialog benefits from the natural progression of textual semantics and acoustic dynamics across dialogue turns. Bidirectional modeling achieves better Recall@40 and Recall@50, but performs worse than forward modeling at Recall@10 and Recall@20. Since accurate top-ranked retrieval is particularly important in our setting, we retain forward modeling as the default configuration due to its stronger shallow-depth retrieval performance and simpler architecture.

\begin{table*}[t!]
\caption{Recall Performance under Different Dialogue Temporal Modeling Strategies.}
\centering
\resizebox{0.82\linewidth}{!}{
\begin{tabular}{lccccc}
\toprule
\textbf{Systems} & \textbf{Recall@10} & \textbf{Recall@20} & \textbf{Recall@30} & \textbf{Recall@40} & \textbf{Recall@50} \\
\midrule
Reversed Dialogue Modeling 
& 40.69 & 58.01 & 68.40 & 76.19 & 83.55 \\

Randomly Shuffled Dialogue Modeling 
& 37.66 & 57.58 & 66.67 & 73.16 & 80.09 \\

Bidirectional Dialogue Modeling 
& 42.86 & 60.17 & \textbf{72.73} & \textbf{82.25} & \textbf{86.15} \\

\midrule
\textbf{S2Dialog (Forward Modeling)} 
& \textbf{50.68} & \textbf{63.01} & 72.60 & 77.17 & 83.56 \\
\bottomrule
\end{tabular}
}
\label{tab:temporal}
\end{table*}

\subsection{Analysis of K Selection}

\begin{figure}
    \centering
    \includegraphics[width=0.85\linewidth]{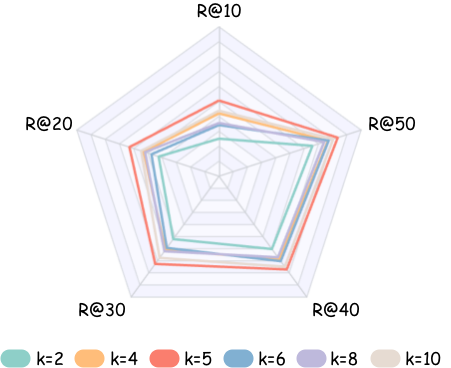}
    \caption{Retrieval performance across various $K$ values. The performance follows an inverted U-shaped trend, peaking at $K=5$ for all Recall metrics. This highlights the crucial role of an appropriate sample size in balancing training sufficiency and data redundancy during the optimization process.}
    \label{fig:k}
\end{figure}

To analyze the impact of different K values on retrieval performance, we train S2Dialog with varying K. As shown in Fig. \ref{fig:k}, when K=2, the model achieves the worst performance across all metrics from R@10 to R@50. As K increases, the performance gradually improves and reaches its peak at K=5. However, when K exceeds 5, the performance starts to decline. These results indicate that an appropriate K value plays a crucial role in the performance of S2Dialog: a too small K leads to insufficient training samples, whereas a too large K introduces redundant information, both of which negatively affect the final performance. Based on this analysis, we adopt K=5 as the default setting in subsequent experiments.

\subsection{Case Study}
\begin{figure}
    \centering
    \includegraphics[width=1\linewidth]{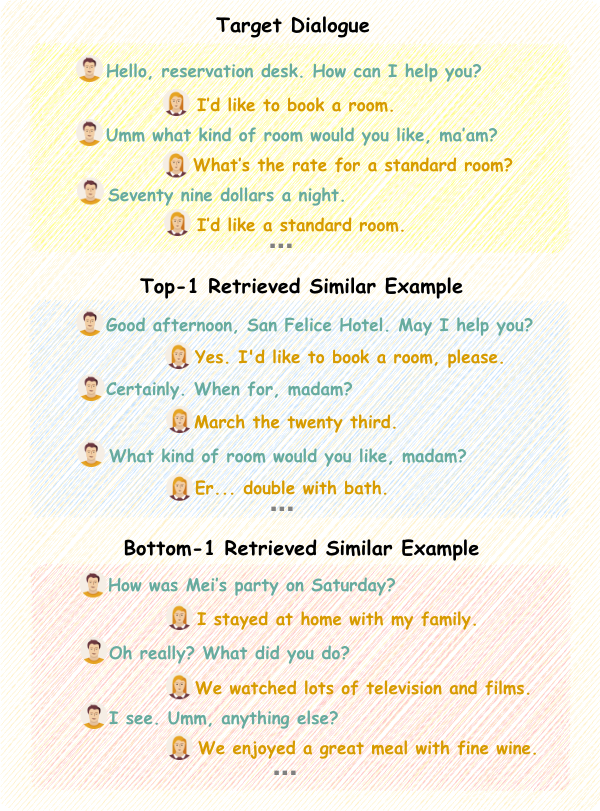}
    \caption{Case study of the S2Dialog.}
    \label{fig:case}
\end{figure}

To provide a more intuitive validation of the scenario similarity between the dialogues retrieved through S2Dialog and the target dialogue, we conducted a case study. As shown in Fig. \ref{fig:case}, the scenario of the target dialogue is a hotel reservation. The target dialogue includes two interlocutors communicating about room types and price confirmation, providing a clear semantic context for retrieval. Due to space limitations, this work presents the Top-1 and Bottom-1 dialogue cases. The Top-1 dialogue scenario is highly consistent with the target dialogue in semantic context, and it covers relevant exchanges such as room type selection. This indicates that S2Dialog can accurately understand the context of the target dialogue and correctly retrieve dialogues that are similar in semantics and style. In addition, the Bottom-1 corresponds to an unrelated daily dialogue scenario, whose semantic context shows a clear difference from hotel reservation. In summary, this case study verifies the effectiveness of S2Dialog in capturing semantic relationships between dialogues.

\section{Conclusion and Future Work}
In this study, we introduced \textit{S2Dialog}, a unified framework that shifts the paradigm from conventional utterance-level aggregation to holistic dialogue-level multimodal modeling. The framework composed of a dialogue-level textual retriever and a dialogue-level acoustic retriever, jointly optimized via dialogue-level textual-acoustic contrastive learning. Experimental validations on the DailyTalk corpus reveal that this synergy significantly enhances the discriminative power between heterogeneous conversational contexts, consistently outperforming representative baselines across all retrieval metrics. These findings underscore the necessity of modeling entire dialogue sequences to retrieve the long-range dependencies and stylistic consistency inherent in human communication.  In the future, we plan to further explore finer-grained dialogue retrieval approaches and extend our system to multilingual or open-domain dialogue scenarios, aiming to improve its generalization and practical applicability.

\section{Limitations}
One limitation of our work is that S2Dialog is currently evaluated on the DailyTalk dataset to validate the effectiveness of dialogue-level semantic-style retrieval. Although DailyTalk provides high-quality multimodal dialogue data with aligned textual and acoustic information, it represents a relatively constrained conversational setting. Future work will extend the evaluation to larger and more diverse multimodal dialogue corpora with more speakers and broader dialogue scenarios. In addition, our current implementation adopts Sentence-BERT and Wav2Vec2-IEMOCAP as frozen utterance-level feature extractors, allowing us to focus on dialogue-level retrieval modeling. Further improvements may be achieved by incorporating stronger pretrained language and speech models or by jointly fine-tuning modality-specific encoders.


\bibliography{cas-refs}

\end{sloppypar}
\end{document}